# Feature Selection Using the Concept of Peafowl Mating in IDS

Partha Ghosh[1], Joy Sharma[2] and Nilesh Pandey[3]

[1, 3]Dept. of Information Technology, Netaji Subhash Engineering College, Kolkata, India
[2]Dept. of Computer Science and Engineering, Netaji Subhash Engineering College, Kolkata, India

## Abstract

*Cloud computing has high applicability as an Internet based service that relies on sharing computing resources. Cloud computing provides services that are Infrastructure based, Platform based and Software based. The popularity of this technology is due to its superb performance, high level of computing ability, low cost of services, scalability, availability and flexibility. The obtainability and openness of data in cloud environment make it vulnerable to the world of cyber-attacks. To detect the attacks Intrusion Detection System is used, that can identify the attacks and ensure information security. Such a coherent and proficient Intrusion Detection System is proposed in this paper to achieve higher certainty levels regarding safety in cloud environment. In this paper, the mating behavior of peafowl is incorporated into an optimization algorithm which in turn is used as a feature selection algorithm. The algorithm is used to reduce the huge size of cloud data so that the IDS can work efficiently on the cloud to detect intrusions. The proposed model has been experimented with NSL-KDD dataset as well as Kyoto dataset and have proved to be a better as well as an efficient IDS.*

## Keywords

*Cloud Computing, Intrusion Detection System (IDS), Feature Selection (FS), Peafowl Mating (PFM) Algorithm, NSL-KDD Dataset, Kyoto Dataset*

## 1. Introduction

Cloud Computing provides the opportunities of accessing the remote activities which are services like platform, software or infrastructure as per the needs over the Internet. It provides the facilities over the Internet which can be accessed from anywhere the user wishes to use. Since the use of this technology is increasing day by day, the privacy of data has become a matter of concern as it attracts more intruders who try to access data illegally [1]. For privacy, IDS provides us with a solution. IDS performs a crucial role in minimizing threats to the information system and maintaining security in the network [2]. IDSs are of mainly two types: Network based IDS (NIDS) and Host based IDS (HIDS). NIDS is located at critical points of the network to investigate all data packets passing across [3]. It analyses the passing packets on the entire subdivision of the network. In case it notices any malicious activity by matching it with previous attack records, it notifies the host and prevents the attack. It has a strong mechanism but with very low visibility in the host server. HIDS is placed on individual hosts. It holds a record of important system files and keeps on checking each incoming and outgoing file. In case any alterations in the records are found, it notifies the administrator. HIDS is not capable of monitoring applications, only the use of resources of the applications can be checked by it. Depending on the finding mechanism of IDS, it can be categorized into misuse detection and anomaly detection [4]. Only the attacks for which a previous record is already present can be detected by misuse detection. Misuse detection is also referred to as signature-based detection and pattern matching. Anomaly detection is implemented to spot novel types of attacks. It can identify any action that is quite different from standard behavior i.e., it can recognize anomalies [5, 6]. Various algorithms are being used to increase the efficiency of IDS. IDS has to deal with a huge amount of data. The datasets are so large that it takes a long time to be processed. These datasets contain many irrelevant data which is not necessary for computation increases time consumption as well as may cause misclassification. This is the reason why the size of the datasets is needed to be reduced [7].

To shorten the training time and memory allocation of the dataset, some of the dimensions of the dataset are dropped without compromising with the efficiency of the system using different kinds of Feature Selection (FS) methods [8, 9]. A large dimensional dataset reduces the effectiveness of the results. So here, a new nature inspired algorithm, Peafowl Mating (PFM) algorithm, is proposed to select relevant Feature Subsets (FSs). After the reduction of features and selection of relevant FSs, the data records are classified either an anomaly or normal record. This provides with better results and leads to a highly effective as well as robust IDS.

The remaining part of the paper is arranged accordingly: Section 2 upholds the pieces of literature on IDS, cloud and so on from which the authors of this paper gathered motivation. Section 3 is the prime part of the paper, demonstrating the novel proposed IDS model. The experimental results of the proposed model are portrayed in details in Section 4. Ultimately, the concluding statements are provided in Section 5.

## 2. RELATED WORKS

Network security has been a matter of concern since the popularization of Internet. Firewalls look into the ways of protecting the devices and systems in various possible methods from attacks. But to increase the rate of detecting the intrusions properly an IDS is needed. T.N. Kim et al. in 2022 suggested a novel approach by combining IDS with firewall which updated the firewall filtering rule if any new type of intrusion was identified [10]. In 2018 Elham Besharati et al. proposed a HIDS which would search and select important features for each class using logistic regression and also used regularization techniques to improve the values. All the attacks are classified using the combination of three popular classifiers. The model was tested on NSL-KDD data set and showed an accuracy of 97.51% for detecting attacks against normal states [11]. In 2022, T.T. Huynh and H.T. Nguyen developed an IDS which used a combination of multilayer Neural Network (NN) with Dense Sparse Dense (DSD) multi-stage training [12]. That model was experimented using Recurrent Neural Network (RNN), Long-Short Term Memory (LSTM), Gated Recurrent Unit (GRU) etc. M.R. Gauthama Raman et al. [13] had developed a proficient IDS. To set parameters, hypergraph based Genetic Algorithm and Support Vector Machine was used and applied for selecting features. To build the IDS with high Detection Rate and low False Positive Rate they also introduced a weighted objective function. In 2004, Srilatha Chebrolu et al. [14] proposed a hybrid IDS by combining Markov Blanket model and decision tree as FS methods. The performance of their model was judged by Bayesian Network (BN) and Regression Tree (CART). Distributed Denial of Service (DDoS) continuously launches attacks on cloud services and makes resources unavailable. In 2018, Mustapha Belouch et al. [15] proposed a Hybrid Filter-Wrapper Feature Selection HFWFS method for DDoS detection. That uses both filter and wrapper methods to detect the most redundant features and generate a reduced feature set. The wrapper method is implemented to achieve the optimal selection of features. For evaluating how it performs, they have used NSL-KDD and UNSW-NB15 datasets and then applied Random Tree classifier. In 2012, Li-Fei Chen et al. [16] proposed a meta-heuristic algorithm to solve the feature selection problem efficiently in a high-dimensional feature space. The improved particle swarm optimization (IPSO) algorithm proposed by them, uses the opposite sign test (OST) which diversifies the population in PSO and improves the jump ability of flying particles so that local optimal trapping can be avoided. The performance is then evaluated on the basis of classification accuracy. In 2018, Gursel Serpen et al. [17] designed a host-based IDS for detecting misuse on Linux operating system which uses a feature extraction technique based on PCA of operating system call trace data and employs k-nearest neighbor algorithm for classification purpose. The propose method was tested on the ADFA-LD dataset which contains six types of attack along with normal records. Eduardo de la Hoz et al. built a classifier using Support Vector Classifier Ensemble. They paid extra attention to the data pre-processing and feature selection. They trained each classifier with a distinct feature set in order to hike the detection abilities for a specific class. They have used linear and non-linear feature selection technique [18]. Seyed Mojtaba Hosseini Bamakan et al. worked on creating an IDS framework using chaos particle swarm optimization. For this work, they modified the chaos particle swarm optimization using the time-varying inertia weight factor (TVIW) and time-varying acceleration coefficients (TVAC). After creating the framework, they selected the subset of features for Multiple Criteria Linear Programming (MCLP) and SVM [19]. It was found that the huge network traffic dataset causes a lot of difficulties to process the dataset. Due to this the accuracy rate and the detection rate decreased in the IDS. In 2015, Raman Singh et al. presented an Online Sequential Extreme Learning Machine (OS-ELM) based IDS with network traffic profiling [20]. In their work, they used an ensemble of three FS techniques to minimize the feature set of network traffic dataset and for training the dataset, Beta profiling had been used. In 2018, Qusay M. Alzubi et al. [21] developed a Modified Binary Grey Wolf Optimization (MBGWO) based intrusion detection system. They split up the work into three different parts. To prove the efficiency that model was compared with a number of existing algorithms. The authors analyzed the mentioned papers to understand how IDS actually works as well as find out its strengths and weaknesses. Inspired by these works the authors have proposed a proficient IDS model to solve the issues related to security on cloud.

## 3. PROPOSED MODEL

In the modern era, protecting the devices and services on the Cloud environment is a great challenge. Day by day attackers are improving their ways to infiltrate into systems and cause disruptions. There is an urgent need to upgrade the security mechanisms to prevent the intrusions. IDS is one of the mechanisms which helps in preserving security on Cloud environment. An IDS scrutinizes data packets to distinguish between normal traffic and attack packets. But there are numerous data packets in the network as well as the packets contains quite a number of attributes. To select out only the useful data packets and extract the most useful features different machine learning techniques need to be implemented. The authors of this paper have worked towards making Cloud a safer place by proposing a novel feature selection method and using it in IDS.

In this article, a nature inspired method to solve the real-world optimization problems have been used by the authors. Nature inspired algorithms are the algorithms in which the behaviors of existing species are observed carefully and algorithms are developed based on those behaviors. Basically, the biologically inspired algorithms are influenced by the natural phenomena [22]. The rules of nature are visualized into algorithms for prolific performance. These algorithms can be used in several machine learning techniques like classification, clustering, data mining, feature selection etc. The complexity of the real-world optimization problems is too high to process in an acceptable amount of time. This is why the algorithms should be such that those can process to solutions in affordable complexity [23]. Here, the authors have been motivated by the mating behavior of Peafowl. This behavior has been introduced into the paper in the form of a nature inspired algorithm. This nature inspired Peafowl Mating based Feature Selection Algorithm provides a near optimal solution to the problem i.e., it's metaheuristic in nature.

For the purpose of the experiment, the authors have used both NSL-KDD and Kyoto datasets. These datasets are needed to be pre-processed and normalized. These contain many redundant values and sometimes the values are not in the form to be directly used for training and testing IDS. So those values need to be in proper numerical form for the purpose to serve. After pre-processing and normalization, the datasets are applied on the proposed model [24]. The PFM Algorithm is then applied over the training data to reduce the dimensions of the data by deleting the irrelevant features. The algorithm based on the mating behavior of Peafowl generates feature subsets out of the complete dataset. After that, the testing dataset is used for classification using different classifiers. Figure 1 depicts the progress of the proposed model.

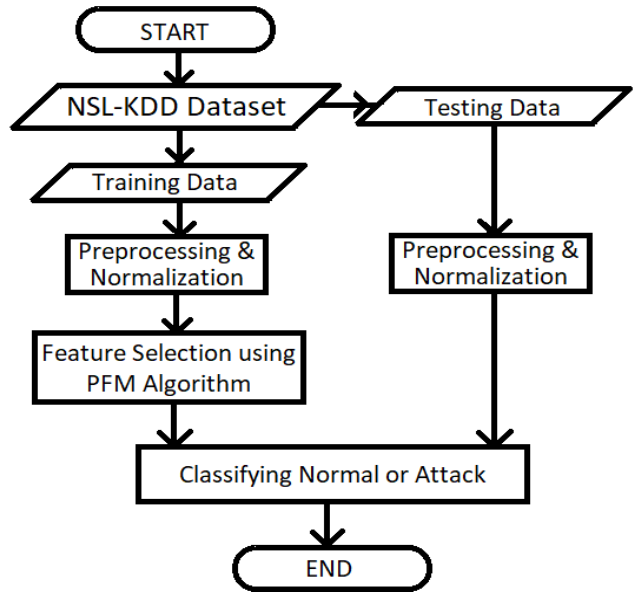

Figure 1. Flowchart of the proposed model

### 3.1. Behavior of Peafowls

By the term Peafowl, both peacock and peahen are addressed. Peacocks elaborate their colorful feathers to attract the peahens. In the mating season, the peacocks spread their feathers to show the dominance and to prove themselves superior compared to the others. The prettier the peacock, the more peahens are attracted by it. A peacock shakes his feathers and struts around with confidence as if he owns the place. Tail feathers of a peacock can be seen underneath the feathers which are spread to attract peahens. Peacocks have also a wide vocabulary of calls, and during the mating season they make a distinctive hoot to attract the peahens. The peahens are attracted by the colorfulness as well as the intensity of the sound or mating call generated by the peacocks. The peahens roam around the territories, after that, they select the peacock to mate with. The peacocks with the higher colorfulness and sound intensity are likely to be chosen by the peahens with higher efficiency. Peacocks are polygamous by nature i.e., they wish to mate with as many peahens as possible. There are some peacocks who are more attractive as well as stronger than others and are likely to mate with multiple peahens in a single season. These types of peacocks are called dominant ones. So, they are basically more efficient creatures in their own kingdom. The Peacocks which are not so efficient are called non-dominant peacocks. Those are less likely to mate with more than one peahen. After mating, peahens leave the place with the fertilized eggs to lay. In this way the generation continues.

## 3.2. Proposed Peafowl Mating Algorithm

In this article, the authors used the behavior of peafowls during mating to implement a model for FS. The model comprises of the random data points which represent the peafowls and some parameters based on which the mating occurs. Some calculation formulae among the data points have been introduced by the authors according to the structure of mating. The new generation is created through interactions of data points as per the behavioral nature of peafowls. To increase the performance of the algorithm the authors assume a number of mating seasons for a single generation. Authors also assume different number of males and females for different seasons. The process continues to a particular terminating condition where the authors have got the results of selected Feature Subsets (FSs). In this model, some key principles have been considered about the peafowl in implementing the algorithm. These are:

- All the peafowls are interpreted as unisex at first. In every generation, according to their fitness values, the peacocks and peahens are differentiated.
- Colorfulness of feathers and sound intensity are key characteristics for the purpose of peafowl mating algorithm.
- In every generation the total number of peafowls are constant.

The algorithm is basically based on two key characteristics of peafowl: the intensity of mating call of the peacock and the colorfulness of their feathers. The attractiveness ($A_i$) of a peafowl $i$, depends on the intensity of mating call $I_0$, the colorfulness of the feathers $C_0$ and the distance $d_{ij}$ between peafowl $i$ and $j$. $A_i$ varies with the change in $d_{ij}$. As the medium is interpreted as consistent, attractiveness varies monotonically and exponentially. The attractiveness is described as

$$A_i = I_0 \cdot e^{-\gamma_1 \cdot d^{ij}} + C_0 \cdot e^{-\gamma_2 \cdot d^{ij}} \tag{1}$$

where, $\gamma_1$ and $\gamma_2$ are sound wave distortion coefficient and color absorption coefficient respectively. Values for $\gamma_1$ and $\gamma_2$ are given as $\gamma_1, \gamma_2 \in [0, +\infty)$. The value of $d_{ij}$ for two peafowls $i$ and $j$ is basically the Euclidean distance between co-ordinates $x_i$ and $x_j$ which can be given as

$$d^{ij} = |x_i - x_j| \tag{2}$$

In this proposed PFM algorithm, the authors have taken n data points as $N$ (set of $n$ number of data points or peafowls) randomly with binary values i.e., 0's and 1's for the experimental purpose. For the first iteration using classifier, the fitness value of each data point is calculated. A random number $r$ is generated (which depends on the size of the population) to differentiate between male ($N_p$) and female ($N_h$). Equation (3) and (4) give the number of males $n_p$ and number of females $n_h$.

$$n_p = n \times r \tag{3}$$
$$n_h = n - n_p \tag{4}$$

To extract $N_p$ and $N_h$ separately, the data points are sorted in descending order according to their fitness value. The top $n_p$ data points are separated as male in $N_p$ and the rest $n_h$ points are stored as female in $N_h$.

There are two types of males, one is normal another one is dominant. The dominant ones are basically capable to mate with multiple females i.e., they much more efficient than the normal

International Journal of Computer Networks & Communications (IJCNC) Vol.16, No.1, January 2024

ones. This type of males is supposed to have higher fitness values. Another random number, Dominance Factor α is generated to indicate the number of dominant males. Value of α is also dependent on population size. Using (5) and (6) the number of dominant males and normal males are calculated.

$$n_d = n_p \times \alpha \quad (5)$$
$$n_n = n_p - n_d \quad (6)$$

Here, $n_d$ is the number of dominant males and $n_n$ is the number of the remaining normal males. The top $n_d$ number of males are denoted by $N_d$ and $N_n$ represents $n_n$ number of other males.

For the mating purpose, the peahens with higher fitness values choose the dominant peacocks which are also of higher fitness values. Every peahen chooses one peacock of its choice. The dominant peacocks can be chosen by more than one peahen. After mating, the newborn peafowl are stored in a set called *new*. A newborn peafowl $k$, born with the peacock $x_i$ and the peahen $x_j$, can be described as

$$new_k = x_i.x_j + (x_i - x_j).\left(I_0.e^{-\gamma_1.d^{ij}} + C_0.e^{-\gamma_2.d^{ij}}\right) + rand.e^{\gamma_1\gamma_2} \quad (7)$$

where, *rand* is a random number generated for including a mutated behavior. The value of *rand* should be a fraction to properly regulate the positional value of the newborn. The value of *rand* lies between -1 to 1 (both inclusive). Some of the features of a newborn is acquired on its own due to evolution and it occurs because of the *rand* variable.

Once the set of newborn peafowl is calculated as *new*, the value of $new_{kd}$ (i.e., $d^{th}$ dimension of the $k^{th}$ data point) can be either in the range of 0 to 1 or out of it. There is a need to transform the real values into binary i.e., 0 or 1 as the selected features are needed to be extracted. So, a probabilistic rule which is based on hyperbolic tangent function is applied to each dimension of the position vector. Equation (8) gives the formula of the function.

$$x_{id} = \begin{cases} 1, & \text{if } rand < S(x_{id}) \\ 0, & \text{otherwise} \end{cases} \quad (8)$$

$$S(x_{id}) = \tanh(|x_{id}|) = \frac{e^{2|x_{id}|} - 1}{e^{2|x_{id}|} + 1} \quad (9)$$

After that, the new peafowls are added to the parent set with their respective fitness values. Now, all the peafowls are sorted in a decreasing manner with accordance to their fitness values. The topmost $n$ number of peafowls are selected for upcoming iteration. In this way, the process continues updating as well as adding the new peafowls and fetching out top $n$ number of peafowls every time. This is done to obtain only the peafowls with higher fitness values and keep the size of the population constant. Finally, after completing all iterations, the peafowl having maximum fitness value is selected as the optimal feature subset. The experiment with the proposed model showed that it has performed better for the work of feature selection in terms of both decreasing the processing power and the memory requirement.

In the proposed model, $\gamma_1$ and $\gamma_2$ are set to 1. $r$ is a random value lying in the range [0.4, 0.6]. Both intensity of mating call $I_0$ and colorfulness of the feathers $C_0$ are set to 0.1 and α is the dominance factor which is regulated at 0.8.

International Journal of Computer Networks & Communications (IJCNC) Vol.16, No.1, January 2024

The pseudo-code of the proposed algorithm is given below:

Algorithm: Feature Selection based on PFM Algorithm

Input: Dataset X which contains x number of records

Output: Optimal Feature Subset

Steps:
```
    1: begin
    2: Generate population N with n number of datapoints
    3: Set values for parameters r, α, γ₁, γ₂
    4: Calculate fitness of each peafowl
    5: Sort them in decreasing order with respect to their fitness values
    6: while (iteration < maximum number of iteration): do
    7:          while (season_number < maximum number of season): do
    8:                  Compute N_p using (3)
    9:                  Compute N_h using (4)
   10:                  Compute N_d using (5)
   11:                  Compute N_n using (6)
   12:                  for (each peacock in N_d): do
   13:                          Mate with multiple peahens in N_h randomly using (7)
   14:                          Store the newborn peafowls
   15:                  end for
   16:                  for (each peacock in N_n): do
   17:                          Mate with single peahen in N_h randomly using (7)
   18:                          Store the newborn peafowls
   19:                  end for
   20:                  Calculate fitness of the newly generated peafowls
   21:                  Append all new peafowls to the population
   22:                  Sort population in decreasing order in accordance to fitness
   23:                  Keep top n peafowls for upcoming iteration
   24:                  season_number = season_number + 1
   25:          end while
   26:          iteration = iteration + 1
   27: end while
   28: end
```

## 4. EXPERIMENTAL RESULTS AND ANALYSIS

For conducting the experiments, scikit-learn library of version 0.20.2 written in Python has been used [25]. The experiments were conducted in an Intel Core i7 7th Gen processor @2.20 GHz system with 8 GB RAM running on Windows 10 Operating System. The environment was set up in a Cloud simulating framework named CloudSim.

### 4.1. Dataset

In 2009, M. Tavallaee et al. proposed NSL-KDD dataset which is a benchmark dataset and is often used for identifying intrusion [26]. This dataset is derived from KDDCUP'99 dataset. The dataset consists of 4 components, 'KDD Train+', 'KDD Test+', '20% KDD Training' and 'KDD Test-21'. In this paper for measuring the performance of the proposed IDS model, KDD Train+ is International Journal of Computer Networks & Communications (IJCNC) Vol.16, No.1, January 2024used for training and KDD Test+ is used for testing. KDD Train+ contains 1,25,973 records and KDD Test+ contains 22,544 records. Each record consists of 41 features, labelled as normal or specific attack types [27, 28]. Figure 2 shows the statistical records in NSL-KDD dataset. Both of the train and test dataset contains the signature of various types of attacks, which are categorized into four types – Denial of Service (DoS), Probe, User to Root (U2R), Remote to Local (R2L).

| NSL-KDD Dataset | DoS | Probe | R2L | U2R | Normal | Total records |
|---|---|---|---|---|---|---|
| KDDTrain+ | 45927 | 11656 | 995 | 52 | 67343 | 125973 |
| 20%KDDTraining+ | 9234 | 2289 | 209 | 11 | 13449 | 25192 |
| KDDTest+ | 7458 | 2421 | 2754 | 200 | 9711 | 22544 |
| KDDTest-21 | 4342 | 2402 | 2754 | 200 | 2152 | 11850 |

Figure 2. Statistical records in NSL-KDD dataset

Another benchmark dataset named Kyoto dataset is used for experiments. Network traffic, from November 1$^{st}$, 2006 to December 31$^{st}$, 2015, is captured by the Kyoto University [29]. To conduct experiments authors have collected a part of the benchmark dataset. Network traffic of March 30$^{th}$, 2010 is gathered and after deleting all duplicate and redundant records, only 1,63,752 connections are taken for experimental purposes. Kyoto dataset contains 16 conditional features and 1 decision making feature [20]. The portion of Kyoto dataset used in this experiment contains 74,837 numbers of normal and 88,915 numbers of attack instances.

Both of the above benchmark datasets contain numeric as well as non-numeric conditional attributes. To transform these non-numerical values into numerical and for making the input suitable, preprocessing is done before the experiments. All non-numerical conditional features of both datasets are assigned with numerical values according to the number of occurrences of those particular feature values. Conditional attributes contain discrete and continuous values. Due to the combination of discrete and continuous values, the range of the feature values are not same. To make the range of all the features same and comparable, normalization is done. In the proposed model, to make the datasets normalized, min-max normalization method is used [30]. After performing preprocessing and normalization NSL-KDD dataset as well as Kyoto dataset are used for performing the testing of the proposed model.

**4.1.1. Experiments using Benchmark Functions**

The numerical proficiency of the proposed Peafowl algorithm is tested using 23 classical benchmark functions. To show the superiority of PFM, it is compared with a number of swarm-based optimization algorithms. The population size of 30 and 500 iterations are taken for each algorithm. All benchmark functions [31, 32] are explained in figure 3, 4 and 5.

| Function | Dim | Range | $f_{min}$ |
|---|---|---|---|
| $F_1(x) = \sum_{i=1}^{n} x_i^2$ | 30 | [-100,100] | 0 |
| $F_2(x) = \sum_{i=1}^{n} \|x_i\| + \prod_{i=1}^{n} \|x_i\|$ | 30 | [-10,10] | 0 |
| $F_3(x) = \sum_{i=1}^{n} \left( \sum_{j-1}^{i} x_j \right)^2$ | 30 | [-100,100] | 0 |
| $F_4(x) = \max_i \{\|x_i\|, 1 \leq i \leq n\}$ | 30 | [-100,100] | 0 |
| $F_5(x) = \sum_{i=1}^{n-1} \left[ 100 \left( x_{i+1} - x_i^2 \right)^2 + (x_i - 1)^2 \right]$ | 30 | [-5,10] | 0 |
| $F_6(x) = \sum_{i=1}^{n} ([x_i + 0.5])^2$ | 30 | [-100,100] | 0 |
| $F_7(x) = \sum_{i=1}^{n} i x_i^4 + random[0,1)$ | 30 | [-1.28,1.28] | 0 |

Figure 3. Unimodal benchmark functions

| Function | Dim | Range | $f_{min}$ |
|---|---|---|---|
| $F_8(x) = \sum_{i=1}^{n} -x_i \sin(\sqrt{\|x_i\|})$ | 30 | [-500,500] | $-418.9892 \times Dim$ |
| $F_9(x) = \sum_{i=1}^{n} [x_i^2 - 10\cos(2\pi x_i) + 10]$ | 30 | [-5.12,5.12] | 0 |
| $F_{10}(x) = -20\exp(-0.2\sqrt{\frac{1}{n}\sum_{i=1}^{n} x_i^2}) - \exp\left(\frac{1}{n}\sum_{i=1}^{n}\cos(2\pi x_i)\right) + 20 + e$ | 30 | [-32,32] | 0 |
| $F_{11}(x) = \frac{1}{4000}\sum_{i=1}^{n} x_i^2 - \Pi_{i=1}^{n}\cos\left(\frac{x_i}{\sqrt{i}}\right) + 1$ | 30 | [-600,600] | 0 |
| $F_{12}(x) = \frac{\pi}{n}\left\{10\sin^2(\pi y_1) + \sum_{i=1}^{n-1}(y_i-1)^2\left[1+10\sin^2(\pi y_{i+1})\right] + (y_n-1)^2\right\} + \sum_{i=1}^{n} u(x_i, 10, 100, 4)$ $y_i = 1 + \frac{x_i+1}{4}$ $u(x_i, a, k, m) = \begin{cases} k(x_i - a)^m & x_i > a \\ 0 & -a < x_i < a \\ k(-x_i - a)^m & x_i < -a \end{cases}$ | 30 | [-50,50] | 0 |
| $F_{13}(x) = 0.1\sin^2(3\pi x_1) + 0.1\sum_{i=1}^{n-1}(x_i-1)^2[1+\sin^2(3\pi x_{i+1})] + 0.1(x_n-1)^2[1+\sin^2(2\pi x_n)] + \sum_{i=1}^{n} u(x_i, 5, 100, 4)$ | 30 | [-50,50] | 0 |

Figure 4. Multimodal benchmark functions

| Function | Dim | Range | $f_{min}$ |
|---|---|---|---|
| $F_{14}(x) = \left(\frac{1}{500} + \sum_{j=1}^{25} \frac{1}{j + \sum_{i=1}^{2}(x_i - a_{ij})^6}\right)^{-1}$ | 2 | [-65,65] | 1 |
| $F_{15}(x) = \sum_{i=1}^{11}\left[a_i - \frac{x_1(b_i^2 + b_i x_2)}{b_i^2 + b_i x_3 + x_4}\right]^2$ | 4 | [-5,5] | 0.00030 |
| $F_{16}(x) = 4x_1^2 - 2.1x_1^4 + \frac{1}{3}x_1^6 + x_1 x_2 - 4x_2^2 + 4x_2^4$ | 2 | [-5,5] | -1.0316 |
| $F_{17}(x) = \left(x_2 - \frac{5.1}{4\pi^2}x_1^2 + \frac{5}{\pi}x_1 - 6\right)^2 + 10\left(1 - \frac{1}{8\pi}\right)\cos x_1 + 10$ | 2 | [-5,5] | 0.398 |
| $F_{18}(x) = [1 + (x_1+x_2+1)^2(19-14x_1+3x_1^2-14x_2+6x_1 x_2+3x_2^2)][30+(2x_1-3x_2)^2(18-32x_1+12x_1^2+48x_2-36x_1 x_2+27x_2^2)]$ | 2 | [-2,2] | 3 |
| $F_{19}(x) = -\sum_{i=1}^{4} c_i \exp\left(-\sum_{j=1}^{3} a_{ij}(x_j - p_{ij})^2\right)$ | 3 | [0,1] | -3.86 |
| $F_{20}(x) = -\sum_{i=1}^{4} c_i \exp\left(-\sum_{j=1}^{6} a_{ij}(x_j - p_{ij})^2\right)$ | 6 | [0,1] | -3.32 |
| $F_{21}(x) = -\sum_{i=1}^{5}\left[(X - a_i)(X - a_i)^T + c_i\right]^{-1}$ | 4 | [0,10] | -10.1532 |
| $F_{22}(x) = -\sum_{i=1}^{7}\left[(X - a_i)(X - a_i)^T + c_i\right]^{-1}$ | 4 | [0,10] | -10.4028 |
| $F_{23}(x) = -\sum_{i=1}^{10}\left[(X - a_i)(X - a_i)^T + c_i\right]^{-1}$ | 4 | [0,10] | -10.5363 |

Figure 5. Fixed-dimensional Multimodal benchmark functions

International Journal of Computer Networks & Communications (IJCNC) Vol.16, No.1, January 2024

The comparison of the optimization results is shown in Table 1. PFM algorithm runs 30 times using new populations generated at random. The average cost function and corresponding standard deviation are used for presenting the results. The PFM algorithm is compared with Whale Optimization Algorithm (WOA) [33], Particle Swarm Optimization (PSO) [34], Gravitational Search Algorithm (GSA) [35]. Results of the mentioned algorithms are mainly taken from [33].

Table 1. Comparison of Optimization Algorithms on Benchmark Functions

| F | PFM | | WOA | | PSO | | GSA | |
|---|---|---|---|---|---|---|---|---|
| | avg | std | avg | std | avg | std | avg | std |
| F1 | 1.6E-18 | 3.89795E-19 | 1.41E-30 | 4.91E-30 | 0.000136 | 0.000202 | 2.53E-16 | 9.67E-17 |
| F2 | 5.36E-10 | 7.10533E-11 | 1.06E-21 | 2.39E-21 | 0.042144 | 0.045421 | 0.055655 | 0.194074 |
| F3 | 8.92E-18 | 2.08747E-18 | 5.39E-07 | 2.93E-06 | 70.12562 | 22.11924 | 896.5347 | 318.9559 |
| F4 | 5.42E-10 | 7.38537E-11 | 0.072581 | 0.39747 | 1.086481 | 0.317039 | 7.35487 | 1.741452 |
| F5 | 31.0265547 | 19.8206229 | 27.86558 | 0.763626 | 96.71832 | 60.11559 | 67.54309 | 62.22534 |
| F6 | 1.62E-18 | 3.5667E-19 | 3.116266 | 0.532429 | 0.000102 | 8.28E-05 | 2.5E-16 | 1.74E-16 |
| F7 | 0.01889554 | 0.00745337 | 0.001425 | 0.001149 | 0.122854 | 0.044957 | 0.089441 | 0.04339 |
| F8 | -8612.22938 | 417.594683 | -5080.76 | 695.7968 | -4841.29 | 1152.814 | -2821.07 | 493.0375 |
| F9 | 66.8610837 | 12.77214487 | 0 | 0 | 46.70423 | 11.62938 | 25.96841 | 7.470068 |
| F10 | 2.74E-10 | 2.71065E-11 | 7.4043 | 9.897572 | 0.276015 | 0.50901 | 0.062087 | 0.23628 |
| F11 | 0.00673594 | 0.008485047 | 0.000289 | 0.001586 | 0.009215 | 0.007724 | 27.70154 | 5.040343 |
| F12 | 6.69E-03 | 0.025871846 | 0.339676 | 0.214864 | 0.006917 | 0.026301 | 1.799617 | 0.95114 |
| F13 | 0.00219747 | 0.004470079 | 1.889015 | 0.266088 | 0.006675 | 0.008907 | 8.899084 | 7.126241 |
| F14 | 1.42328322 | 1.635213484 | 2.111973 | 2.498594 | 3.627168 | 2.560828 | 5.859838 | 3.831299 |
| F15 | 0.00041852 | 0.00013061 | 0.000572 | 0.000324 | 0.000577 | 0.000222 | 0.003673 | 0.001647 |
| F16 | -1.0316285 | 6.77522E-16 | -1.03163 | 4.2E-07 | -1.03163 | 6.25E-16 | -1.03163 | 4.88E-16 |
| F17 | 0.39788736 | 1.1292E-16 | 0.397914 | 2.7E-05 | 0.397887 | 0 | 0.397887 | 0 |
| F18 | 3 | 4.51681E-16 | 3 | 4.22E-15 | 3 | 1.33E-15 | 3 | 4.17E-15 |
| F19 | -3.8627798 | 1.80672E-15 | -3.85616 | 0.002706 | -3.86278 | 2.58E-15 | -3.86278 | 2.29E-15 |
| F20 | -3.203161918 | 2.25841E-15 | -2.98105 | 0.376653 | -3.26634 | 0.060516 | -3.31778 | 0.023081 |
| F21 | -10.05352692 | 1.80672E-15 | -7.04918 | 3.629551 | -6.8651 | 3.019644 | -5.95512 | 3.737079 |
| F22 | -10.0637085 | 0 | -8.18178 | 3.829202 | -8.45653 | 3.087094 | -9.68447 | 2.014088 |
| F23 | -10.07504591 | 3.61345E-15 | -9.34238 | 2.414737 | -9.95291 | 1.782786 | -10.5364 | 2.6E-15 |

The exploitation capability of the proposed algorithm is evaluated using the unimodal benchmark functions F1-F7 because they have only one global optimum. Comparative results show that the PFM algorithm provides very good exploitation. To evaluate the ability of exploration of the optimization algorithm, F8-F23 benchmark functions are considered. F8 to F23 functions are multimodal functions that include multiple local optima which increase exponentially with the size of the problem. Comparative results also show very good exploration capabilities of the proposed meta-heuristic algorithm. Table 1 reflects that the proposed novel PFM algorithm makes a balance between the exploitation as well as exploration during the search and is also successful in finding the optimum solution.

**4.1.2. Experiments using Datasets**

Datasets contain a number of attributes, among which some are useless and if those are not excluded, it takes more time for evaluation as well as the accuracy rate decreases. Therefore, it is required to minimize the dimension of such a dataset by selecting only the relevant features.

Here, a nature inspired PFM algorithm has been used for selecting appropriate feature subset from the NSL-KDD and Kyoto dataset. With this proposed model the authors select subsets of features from the original dataset for achieving better performance. Using the selected subsets of features the authors classified the connections as normal or anomaly. Table 2 and 3 show all the selected feature subsets using the PFM algorithm from different datasets. For the further work of classification, the authors divide their work in two different phases.

Table 2. Feature Subsets of NSL-KDD dataset using PFM algorithm

| Feature Subset (FSs) | No. of features | Features |
|---|---|---|
| FSs1 | 21 | 1,3,4,5,6,7,8,9,10,13,15,19,23,26,32,35,36,37,38,39,40 |
| FSs2 | 19 | 2,3,4,5,6,7,8,9,10,13,18,23,25,28,32,35,36,37,40 |
| FSs3 | 19 | 3,4,5,6,7,8,12,17,20,23,25,27,32,35,36,37,38,40,41 |

Table 3. Feature Subsets of Kyoto dataset using PFM algorithm

| Feature Subset (FSs) | No. of features | Features |
|---|---|---|
| FSs1 | 8 | 4,5,6,9,10,12,14,16 |
| FSs2 | 8 | 2,3,5,6,9,10,14,16 |
| FSs3 | 9 | 1,5,6,9,10,11,13,14,16 |

Scenario 1: In the first scenario only NSL-KDD dataset is taken. Here, authors have taken NSL-KDD train dataset for training and NSL-KDD test dataset to judge the performance of reduced datasets. A number of different classifiers have been applied to measure the performance. Table 4 presents the obtained results.

Performance of any IDS depends on four parameters [36] – True Positive (TP), True Negative (TN), False Negative (FN) and False Positive (FP). Based on these four parameters, performance metrics are calculated which are as follows:

$$\text{Accuracy (AC)} = \frac{TP+TN}{TP+TN+FP+FN} \quad (10)$$
$$\text{Detection Rate (DR)} = \frac{TP}{TP+FN} \quad (11)$$
$$\text{False Positive Rate (FPR)} = \frac{FP}{TN+FP} \quad (12)$$
$$\text{True Negative Rate (TNR)} = \frac{TN}{TN+FP} \quad (13)$$
$$\text{False Negative Rate (FNR)} = \frac{FN}{FN+TP} \quad (14)$$
$$\text{Precision} = \frac{TP}{TP+FP} \quad (15)$$
$$\text{F1score} = \frac{2 \times TP}{2 \times TP+FP+FN} \quad (16)$$

Here, the authors extracted three Feature Subsets (FSs) from the original NSL-KDD dataset with the feature number counts of 21, 19 and 19 features respectively. For each selected FSs, the Accuracy (AC), Detection Rate (DR) and False Positive Rate (FPR) have been calculated with selected features and using all features. It is found that in every case, reduced set of features gave better performance as well as consumed less memory space. Hence, the proposed PFM algorithm creates a robust and efficient IDS. The proposed model was also compared with GWO, MGWO, BGWO and MBGWO algorithms. Results obtained from experiments highlight that the proposed algorithm achieves higher accuracy. The comparison depicted in Table 5 shows that the proposed model outperforms the models developed by other authors. After classification, Table 6 shows all the performance metrics for FSs1. Fig. 6 to Fig. 10 illustrate all the performance metrics for FSs1 using NN, DT, KNN, Bagging and RF respectively.

Table 4. Comparative study of Scenario 1 on NSL-KDD dataset

| Classifier | Feature Subset | With Feature Selection | | | Without Feature Selection | | |
|---|---|---|---|---|---|---|---|
| | | AC (%) | DR (%) | FPR (%) | AC (%) | DR (%) | FPR (%) |
| Neural Network | FSs1 | 82.244 | 70.973 | 2.863 | 79.467 | 66.337 | 3.182 |
| | FSs2 | 81.490 | 69.750 | 2.997 | | | |
| | FSs3 | 83.029 | 72.688 | 3.306 | | | |
| Decision Tree | FSs1 | 84.816 | 75.999 | 3.532 | 79.649 | 70.077 | 7.703 |
| | FSs2 | 83.863 | 74.036 | 3.151 | | | |
| | FSs3 | 82.248 | 71.215 | 3.172 | | | |
| K-Nearest Neighbor | FSs1 | 80.376 | 67.397 | 2.471 | 77.608 | 62.417 | 2.317 |
| | FSs2 | 82.173 | 70.482 | 2.379 | | | |
| | FSs3 | 80.239 | 67.085 | 2.379 | | | |
| Bagging | FSs1 | 83.299 | 73.163 | 3.306 | 80.265 | 68.869 | 4.675 |
| | FSs2 | 83.047 | 72.485 | 2.997 | | | |
| | FSs3 | 82.031 | 70.825 | 3.161 | | | |
| Random Forest | FSs1 | 79.551 | 66.212 | 2.822 | 78.149 | 63.867 | 2.976 |
| | FSs2 | 79.108 | 65.534 | 2.955 | | | |
| | FSs3 | 79.241 | 65.760 | 2.945 | | | |

Table 5. Comparison with other Feature Selection methods on NSL-KDD dataset in Scenario 1

| Algorithm | Average Accuracy (%) | Average number of selected features |
|---|---|---|
| KNN-NN [37] | 76.54 | 25 |
| GWO [21] | 79.66 | 28 |
| MGWO [21] | 79.66 | 24 |
| bGWO [21] | 81.07 | 26 |
| MBGWO [21] | 81.58 | 26 |
| PFM | 81.784 | 20 |

Table 6. Performance metrics of FSs1 of NSL-KDD dataset

| Classifier | Performance Metrics | Without Feature Selection (%) | With Feature Selection (%) |
|---|---|---|---|
| Neural Network | AC | 79.241 | 82.244 |
|  | DR | 65.877 | 70.973 |
|  | FPR | 3.100 | 2.863 |
|  | TNR | 96.900 | 97.137 |
|  | FNR | 34.123 | 29.027 |
|  | Precision | 96.562 | 97.038 |
|  | F1score | 78.321 | 81.984 |
| Decision Tree | AC | 79.666 | 84.816 |
|  | DR | 70.085 | 75.999 |
|  | FPR | 7.672 | 3.532 |
|  | TNR | 92.328 | 96.468 |
|  | FNR | 29.915 | 24.001 |
|  | Precision | 92.350 | 96.603 |
|  | F1score | 79.692 | 85.071 |
| K-Nearest Neighbor | AC | 77.608 | 80.376 |
|  | DR | 62.417 | 67.397 |
|  | FPR | 2.317 | 2.582 |
|  | TNR | 97.683 | 97.529 |
|  | FNR | 37.583 | 32.603 |
|  | Precision | 97.268 | 97.300 |
|  | F1score | 79.039 | 79.634 |
| Bagging | AC | 81.250 | 83.299 |
|  | DR | 70.553 | 73.163 |
|  | FPR | 4.613 | 3.306 |
|  | TNR | 95.387 | 96.694 |
|  | FNR | 29.448 | 26.837 |
|  | Precision | 95.285 | 96.694 |
|  | F1score | 81.075 | 83.299 |
| Random Forest | AC | 77.599 | 79.551 |
|  | DR | 62.893 | 66.212 |
|  | FPR | 2.966 | 2.822 |
|  | TNR | 97.034 | 97.179 |
|  | FNR | 37.108 | 33.788 |
|  | Precision | 96.555 | 96.876 |
|  | F1score | 76.170 | 78.661 |

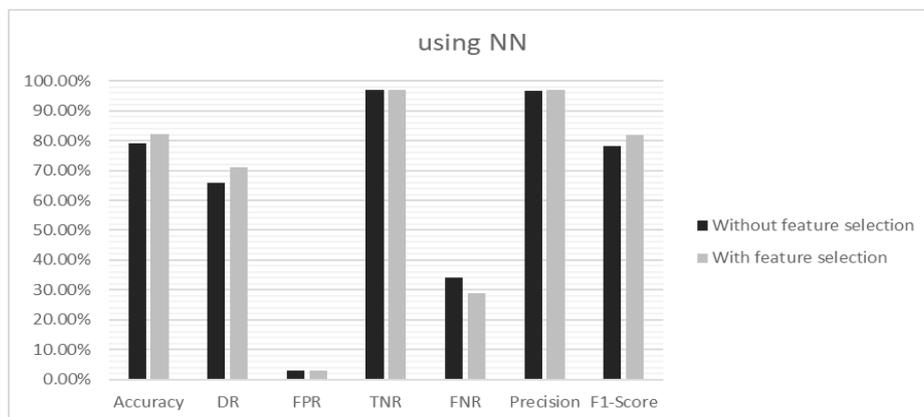

Figure 6. Performance metrics of FSs1 of NSL-KDD dataset using NN in Scenario 1

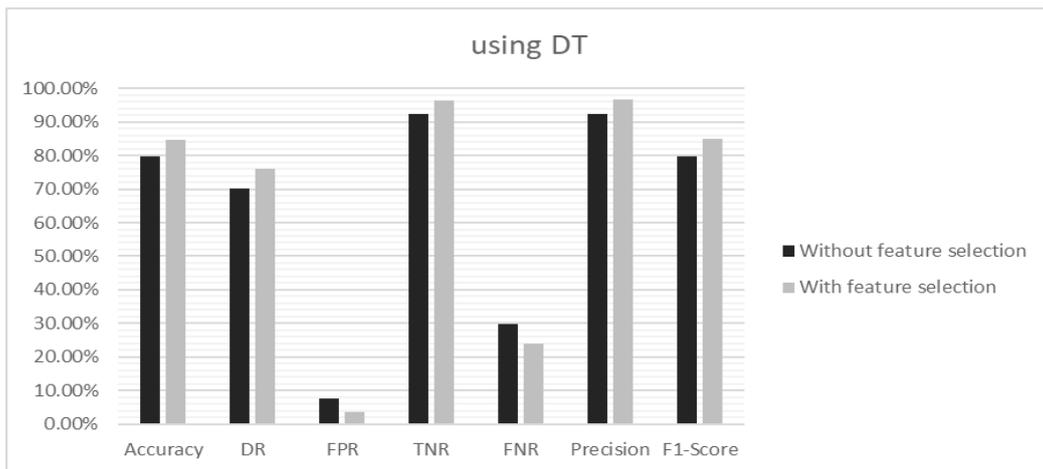

Figure 7. Performance metrics of FSs1 of NSL-KDD dataset using DT in Scenario 1

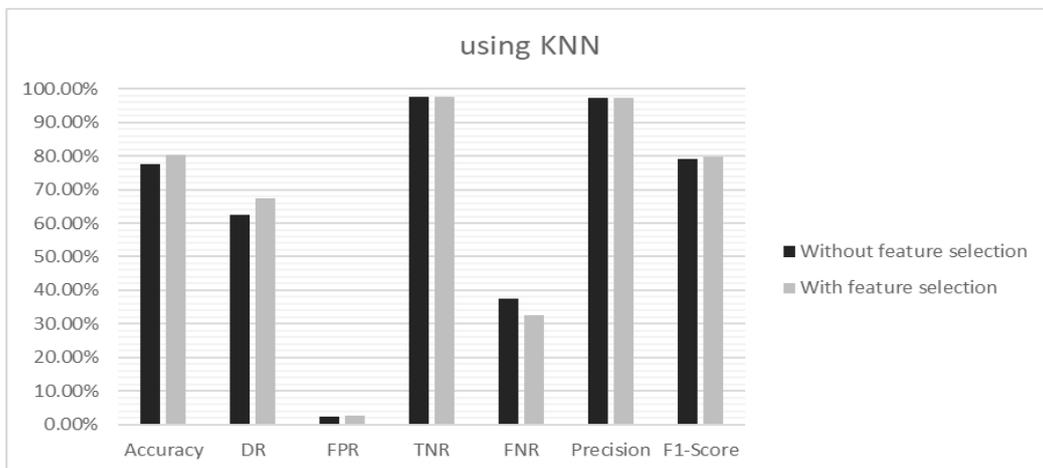

Figure 8. Performance metrics of FSs1 of NSL-KDD dataset using KNN in Scenario 1

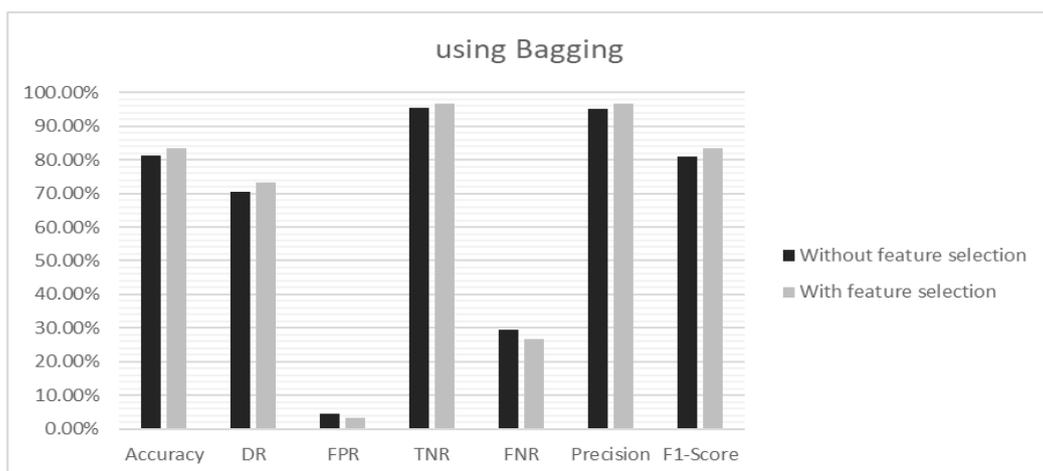

Figure 9. Performance metrics of FSs1 of NSL-KDD dataset using Bagging in Scenario 1

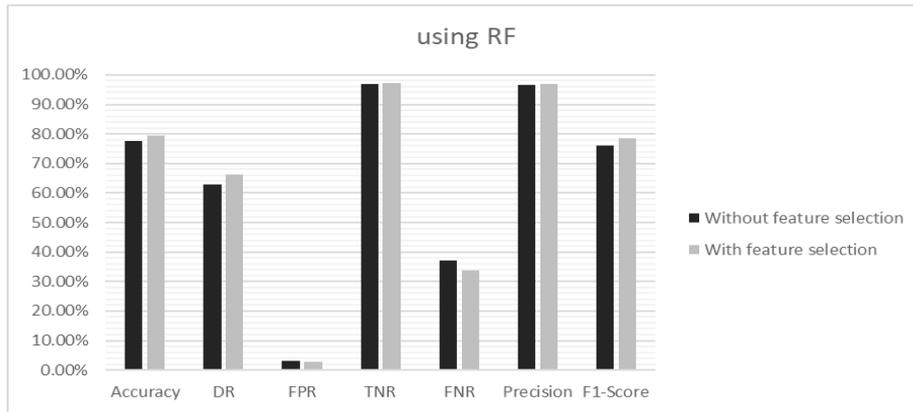

Figure 10. Performance metrics of FSs1 of NSL-KDD dataset using RF in Scenario 1

Scenario 2: In this scenario, both NSL-KDD train set (1,25,973 records) and Kyoto dataset are considered. 10-fold cross validation is applied on the two datasets. It means that a dataset is split into 10 sections or folds. The AC, DR and FPR of the classifiers – NN, DT, KNN, Bagging and RF are noted in Table 7 and 8. The classifiers are trained in two ways – (i) classifiers trained with all features and (ii) classifiers trained with the selected FSs found by the PFM algorithm. The result shows that in most of the cases the selected subset of features gives higher accuracy and better results than the complete dataset.

Apart from this, in this scenario a comparative study featuring the recent advancements in IDS on DR and FPR is shown. This study is presented in Table 9. From these comparisons, it is evident that the proposed IDS model based on Peafowl Mating algorithm performs better than other mentioned IDS models. The PFM algorithm detects intrusions with high efficacy resulting the DR to be 99.568% as well as rarely misjudges normal packets as attacks which generates FPR of 0.271% only on an average.

Table 7. Comparative study of Scenario 2 on NSL-KDD dataset

| Classifier | Feature Subset | With Feature Selection | | | Without Feature Selection | | |
|---|---|---|---|---|---|---|---|
| | | AC (%) | DR (%) | FPR (%) | AC (%) | DR (%) | FPR (%) |
| Neural Network | FSs1 | 99.437 | 99.282 | 0.428 | 99.364 | 99.224 | 0.514 |
| | FSs2 | 99.474 | 99.291 | 0.367 | | | |
| | FSs3 | 99.305 | 99.082 | 0.500 | | | |
| Decision Tree | FSs1 | 99.755 | 99.717 | 0.212 | 99.712 | 99.710 | 0.287 |
| | FSs2 | 99.728 | 99.710 | 0.257 | | | |
| | FSs3 | 99.746 | 99.725 | 0.236 | | | |
| K-Nearest Neighbor | FSs1 | 99.517 | 99.429 | 0.407 | 99.412 | 99.330 | 0.517 |
| | FSs2 | 99.572 | 99.521 | 0.383 | | | |
| | FSs3 | 99.494 | 99.482 | 0.496 | | | |
| Bagging | FSs1 | 99.802 | 99.736 | 0.141 | 99.785 | 99.731 | 0.168 |
| | FSs2 | 99.779 | 99.724 | 0.174 | | | |
| | FSs3 | 99.796 | 99.748 | 0.162 | | | |
| Random Forest | FSs1 | 99.801 | 99.681 | 0.095 | 99.788 | 99.673 | 0.111 |
| | FSs2 | 99.811 | 99.707 | 0.098 | | | |
| | FSs3 | 99.794 | 99.685 | 0.110 | | | |

Table 8. Comparative study of Scenario 2 on Kyoto dataset

| Classifier | Feature Subset | With Feature Selection | | | Without Feature Selection | | |
|---|---|---|---|---|---|---|---|
| | | AC (%) | DR (%) | FPR (%) | AC (%) | DR (%) | FPR (%) |
| Neural Network | FSs1 | 99.408 | 99.425 | 0.612 | 99.307 | 99.504 | 0.926 |
| | FSs2 | 99.676 | 99.792 | 0.462 | | | |
| | FSs3 | 99.437 | 99.495 | 0.632 | | | |
| Decision Tree | FSs1 | 99.820 | 99.866 | 0.234 | 99.809 | 99.821 | 0.206 |
| | FSs2 | 99.830 | 99.882 | 0.232 | | | |
| | FSs3 | 99.819 | 99.828 | 0.192 | | | |
| K-Nearest Neighbor | FSs1 | 98.748 | 98.969 | 1.514 | 98.211 | 98.474 | 2.100 |
| | FSs2 | 98.843 | 99.028 | 1.376 | | | |
| | FSs3 | 98.855 | 99.010 | 1.328 | | | |
| Bagging | FSs1 | 99.836 | 99.881 | 0.218 | 99.831 | 99.850 | 0.191 |
| | FSs2 | 99.837 | 99.885 | 0.220 | | | |
| | FSs3 | 99.841 | 99.865 | 0.187 | | | |
| Random Forest | FSs1 | 99.839 | 99.883 | 0.214 | 99.817 | 99.876 | 0.253 |
| | FSs2 | 99.828 | 99.886 | 0.240 | | | |
| | FSs3 | 99.854 | 99.876 | 0.171 | | | |

Table 9. Comparison with other FS methods on NSL-KDD dataset for Scenario 2

| Authors | Detection Rate (%) | False Alarm Rate (%) |
|---|---|---|
| Singh et al. [20] | 97.67 | 1.74 |
| De la Hoz et al. [18] | 93.40 | 14 |
| Tavallaee et al. [24] | 80.67 | NA |
| Bamakan et al. [19] | 97.03 | 0.87 |
| Raman et al. [12] | 97.14 | 0.83 |
| Abd-Eldayem [38] | 99.03 | 1.0 |
| Kim et al. [39] | 99.10 | 1.2 |
| Gogoi et al. [40] | 98.88 | 1.12 |
| PFM (average) | 99.568 | 0.271 |

# 5. CONCLUSIONS AND FUTURE WORK

Data security is emerging as a great issue in the world of cloud computing. As the data on cloud is placed at remote locations, the authentication of users to access data is becoming a serious threat. This is the time when the data scientists need to work on classifying the authenticated users and the intruders accurately as well as efficiently. An efficient IDS is needed to be developed to suffice this insecure situation. Figuring out the intruders is no less than the ultimate tough thing, as the size of data on Cloud is so large. This is why the algorithm should work in such a manner so that it processes the data most efficiently without compromising with the accuracy. The Peafowl Mating (PFM) Algorithm proposed in this paper has shown better outcomes through Feature Selection approach resulting in increased classification accuracy by reducing the dimensions of the dataset. The authors have shown that the proposed algorithm has performed better than the mentioned ones. As examined by the authors in this paper, the average accuracy of PFM algorithm during train-test scenario is 81.784% which shows the immense potency of this algorithm. The classification results on five different classifiers namely NN, DT, KNN, Bagging and RF show better results and proficiency of the proposed IDS. This proves that the proposed IDS model using PFM algorithm is capable of securing the Cloud environment from attackers. In the future, the proposed Feature Selection algorithm can be calibrated with the classifiers other than the ones used in this paper to perform experiments which may produce even better results. Therefore, an efficient IDS is created that can be deployed in a cloud environment to make it secure enough by detecting the attacks of the intruders which is discussed in this paper; forwarding to more secure and stable system with surpassed accuracy and efficiency.

**CONFLICT OF INTEREST**

The authors declare no conflict of interest.

International Journal of Computer Networks & Communications (IJCNC) Vol.16, No.1, January 202468

## AUTHORS

**Partha Ghosh** achieved his Ph.D. degree in Computer Science & Engineering from Maulana Abul Kalam Azad University of Technology (MAKAUT), West Bengal, India in 2023. He completed B.Sc. in Computer Science (Hons.), M.Sc. in Computer & Information Science and M.Tech. in Computer Science & Engineering from University of Calcutta in 1999, 2001 and 2003 respectively. He has been working as an Assistant Professor at Netaji Subhash Engineering College, Kolkata, West Bengal, India since 2003. His research interests are Cloud Computing, Machine Learning, Intrusion Detection System, Optimization Technique, Feature Selection, Computer Networks and Security etc.

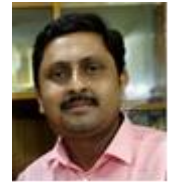

**Joy Sharma** completed his B.Tech. in Computer Science and Engineering from Netaji Subhash Engineering College (affiliated to MAKAUT) in 2019. He is currently working as Software Development Engineer II at Amazon Development Centre (India) Private. He is interested to work in Data Mining, Machine Learning, Artificial Intelligence, Cloud Computing.

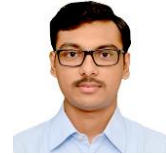

**Nilesh Pandey** completed his B.Tech. in Information Technology from Netaji Subhash Engineering College (affiliated to MAKAUT) in 2020. He is currently working as Software Development Engineer II at Zinier. His areas of interests and experience include Machine Learning, Cloud Computing and Web-based Applications. He has worked on developing intelligent Intrusion Detection systems during his bachelor's studies.

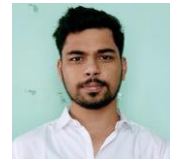